\documentclass{article}

\usepackage{microtype}
\usepackage{graphicx}
\usepackage{subfigure}
\usepackage{booktabs} 

\usepackage{hyperref}

\def\bm{{\bf m}}

\def\0{{\bf 0}}
\def\1{{\bf 1}}

\def\etal{\emph{et al. }}
\def\ie{\emph{i.e.}}
\def\eg{\emph{e.g.}}

\usepackage{times}
\usepackage{epsfig}
\usepackage{amsmath}
\usepackage{amssymb}

\usepackage{multirow}
\usepackage{gensymb}
\usepackage{bm}
\usepackage{xspace}
\usepackage{diagbox}
\usepackage{makecell}
\usepackage[table,dvipsnames]{xcolor}
\definecolor{weijian}{gray}{.9}

\usepackage[accepted]{icml2021}

\icmltitlerunning{What Does Rotation Prediction Tell Us about Classifier Accuracy under Varying Testing Environments?}

\begin{document}

\twocolumn[
\icmltitle{What Does Rotation Prediction Tell Us about Classifier Accuracy \\under Varying Testing Environments?}



\icmlsetsymbol{equal}{*}
\begin{icmlauthorlist}
\icmlauthor{Weijian Deng}{anu}
\icmlauthor{Stephen Gould}{anu}
\icmlauthor{Liang Zheng}{anu}
\end{icmlauthorlist}

\icmlaffiliation{anu}{College of Engineering and Computer Science, Australian National University, Canberra, ACT 0200, Australia}

\icmlcorrespondingauthor{Liang Zheng}{liang.zheng@anu.edu.au}

\icmlkeywords{Machine Learning, ICML}

\vskip 0.3in
]



\printAffiliationsAndNotice{} 
\begin{abstract}
Understanding classifier decision under novel environments is central to the community, and a common practice is evaluating it on labeled test sets. However, in real-world testing, image annotations are difficult and expensive to obtain, especially when the test environment is changing. A natural question then arises: given a trained classifier, can we evaluate its accuracy on varying unlabeled test sets? In this work, we train semantic classification and rotation prediction in a multi-task way. On a series of datasets, we report an interesting finding, \emph{i.e.}, the semantic classification accuracy exhibits a strong linear relationship with the accuracy of the rotation prediction task (\textcolor{black}{Pearson's Correlation  $r > 0.88$}). This finding allows us to utilize linear regression to estimate classifier performance from the accuracy of rotation prediction which can be obtained on the test set through the freely generated rotation labels. 
\end{abstract}

\section{Introduction}
Knowing the generalization ability of a classifier is important for its applications in fields like autonomous driving, medical diagnosis and product recommendation. Many well-established benchmarks in the community, \emph{e.g.,} ImageNet \cite{deng2009imagenet} and CIFAR \cite{krizhevsky2009learning}, provide held-out labeled test sets that allow empirical results to be computed. 
However, when applying the classifier in real-world scenarios, we often encounter novel distributions and unknown cases, and empirical results calculated from existing benchmarks do not accurately reflect its generalization ability during deployment.
To evaluate classifier performance under novel environments, the standard way is to collect new labeled test sets. However, it is prohibitively expensive to label sufficient images in every novel scenario.

This work investigates an interesting problem: how to estimate classifier accuracy on unlabeled test sets? \textcolor{black}{Some existing research studies classifier generalization on \emph{fixed} test sets by developing complexity measurements on model parameters \cite{arora2018stronger,corneanu2020computing,jiang2018predicting}.
In this work, we focus on \emph{varying test domains}.
In the absence of test labels, we aim to find an unsupervised evaluation criterion that can reflect classifier accuracy. \textcolor{black}{A self-supervised task produces ground truths from data without human annotations. Thus, we can always calculate its accuracy on an unlabeled test set.} As such we raise a question: given a network that is good at a self-supervised task on an unlabeled test set, can it also address object recognition task well under the same environment?
If such a relationship exists, we can infer classifier performance on an unlabeled test set from the accuracy of the self-supervised task. In light of this question, we study the statistical correlation between semantic classification and rotation prediction under varying environments.
}

\textcolor{black}{In the community, self-supervised tasks are usually used to learn effective representations for downstream tasks \cite{he2020momentum,noroozi2016unsupervised,gidaris2018unsupervised}. Different from them, we explore the feasibility of using self-supervision to predict classifier accuracy on novel test sets.} We pay particular attention to the pretext task of \emph{rotation prediction}, where we classify the rotation angle of an image into \{0\degree, 90\degree, 180\degree, 270\degree\}.
Specifically, we train a network jointly for the main classification task and pretext task. To study the accuracy correlation between the two tasks, we construct different setups, including CIFAR-10, MNIST, Tiny-ImageNet, and COCO. Each setup contains many labeled test sets for calculating network performance on the two tasks. By plotting classification accuracy against rotation prediction accuracy, we consistently observe a strong correlation between them (Pearson's Correlation $r >0.88$) across the four setups. 
This shows that if the network can solve the rotation prediction task well, it is likely to achieve good classification accuracy, and vice versa.

Based on this, we learn a linear regression model to predict classification accuracy on unseen test sets. Given an unlabeled test set, we obtain rotation ground truths by manually rotating the images. Then, we use the multi-task network to predict rotation degrees of the test images and calculate rotation prediction accuracy. Finally, the linear regression model predicts the semantic classification performance from rotation prediction accuracy.
Empirical results on three different setups show that our regression model achieves very promising estimation results on unseen test sets.

In summary, this work investigates the problem of unsupervised classifier prediction, \ie, estimating classifier performance on unseen test sets without any human-annotated label. We observe that the accuracy of a simple self-supervised task, rotation prediction \cite{gidaris2018unsupervised}, is highly correlated with the classification accuracy under various testing environments. 

\section{Related Work} \label{sec:related}
\textbf{Model generalization.} Estimating model generalization error on unseen images is a crucial problem. 
Some works develop \emph{complexity measurements} on trained models and training sets to predict the generalization gap between a certain pair of training-testing set \cite{eilertsen2020classifying,unterthiner2020predicting,arora2018stronger,corneanu2020computing,jiang2018predicting,neyshabur2017exploring,jiang2019fantastic}.
The authors of \cite{corneanu2020computing} use persistent topology measures to predict the performance gap between training and testing error. 
Jiang \etal \cite{jiang2018predicting} introduce a measurement of layer-wise margin distributions for generalization ability. Neyshabur \etal \cite{neyshabur2017exploring} develop bounds on the generalization gap based on the product of norms of the weights across layers.
 Our work differs significantly. Instead of studying the statistics of model weights, we use rotation prediction accuracy to characterize models. Moreover, we are concerned with various test distributions that have domain gaps with the training one, while the aforementioned works typically assume a fixed test set without a domain gap. 

In addition, some methods try to predict accuracy on unlabeled test samples based on the agreement score of several classifies' predictions \cite{madani2004co,platanios2016estimating,platanios2017estimating,donmez2010unsupervised}. Platanios \etal \cite{platanios2017estimating} use a probabilistic soft logic (PSL) model to infer the errors of the classifiers. 
Recent work also considers the classifier accuracy estimation and proposes to predict accuracy based on the distribution difference \cite{deng2020labels}.
In this work, we attempt to estimate classifier accuracy on unlabeled test sets by leveraging self-supervision.

\textbf{Self-Supervised learning.} Self-supervised learning aims to learn good representations on unlabeled data for the downstream tasks \cite{he2020momentum,noroozi2016unsupervised,gidaris2018unsupervised,caron2018deep,bojanowski2017unsupervised,kolesnikov2019revisiting,doersch2017multi,lee2019making,ren2018cross}. To achieve this, several pretext tasks are proposed, such as solving jigsaw puzzles \cite{noroozi2016unsupervised}, colorization \cite{larsson2017colorization}, rotation prediction \cite{gidaris2018unsupervised}, denoising auto-encoders \cite{vincent2008extracting}, tracking \cite{Wang_2015_ICCV}, and clustering \cite{caron2019unsupervised, caron2018deep}. 
\textcolor{black}{While these works use self-supervision for learning unsupervised representations, this work investigates its utility for unsupervised accuracy estimation.}

Reed \etal \cite{metzger2020evaluating} use rotation prediction is used to automatically select augmentation policies for unsupervised contrastive learning method (\eg, MoCo \cite{he2020momentum}). Liu \cite{liu2020labels} leverage self-supervised pretext tasks for unsupervised neural architecture search.
Self-supervised methods can also be used to improve supervised classifier learning \cite{hendrycks2019using,sun2020test,carlucci2019domain,sun2019unsupervised}. Hendrycks \etal \cite{hendrycks2019using} show jointly training classification and self-supervised task improves classifier robustness and uncertainty estimation. Sun \etal \cite{sun2020test} achieve test-time training by using self-supervised task on unlabeled test images. Moreover, the supervised tasks can be used to alleviate distribution shifts \cite{carlucci2019domain, sun2019unsupervised,bucci2021self}.
\textcolor{black}{This work does not aim to improve supervised classification by using self-supervised tasks. Instead, we study the feasibility of using the rotation prediction task for estimating classifier accuracy.}

\textbf{Out-of-distribution detection.} The overall goal of out-of-distribution is to detect test samples from a novel distribution (a distribution different from the training distribution) \cite{hsu2020generalized,hendrycks2016baseline,lee2018simple,liang2017enhancing,guo2017calibration,ovadia2019can,ren2019likelihood}. Hendrycks \etal \cite{hendrycks2016baseline} use probability outputs from a softmax classifier as an indicator to find out-of-distribution samples. Liang \etal \cite{liang2017enhancing} use temperature scaling \cite{guo2017calibration} and input pre-processing to improve detection accuracy.
Ren \etal \cite{ren2019likelihood} investigate deep generative model-based methods for the sample detection.
While the above methods aim to capture abnormal samples during the testing, our work focuses on predicting overall accuracy that involves all test samples.

\section{Classifier Accuracy Estimation}
\subsection{Problem Definition}
In the problem of unsupervised accuracy estimation, we are provided with a labeled training set and a set of unlabeled datasets for testing. 
We denote the training set as $\mathcal{D}_{train} = \{(\bm{x}_i,{\bm y}_i)\}_{i=1}^{n}$, where $\bm{x}_i$ is the $i$-th training image, ${\bm y}_i \in \{0, 1, ..., K-1\}$ is its label, and $n$ is the total number of images. 
Similarly, we denote each test dataset as ${{\cal D}_{test}^{t}} = \{ {\bm{x}}_j^t\} _{j = 1}^{{n_{t}}}$, where $t \in \{1,2, ..., M\}$ indicates the $t$-th test set, $\bm{x}_j^t$ is the $j$-th test image, and $n_{t}$ is the total number of images in ${{\cal D}_{test}^{t}}$. 
Given a classifier trained on $\mathcal{D}_{train}$, our goal is to estimate its accuracy on {every unlabeled test set ${{\cal D}_{test}^{t}}, t=1,2,...,M$.}

\subsection{Correlation Study} \label{correlation}

\begin{figure}[t]
\begin{center}
\includegraphics[width=1.0\linewidth]{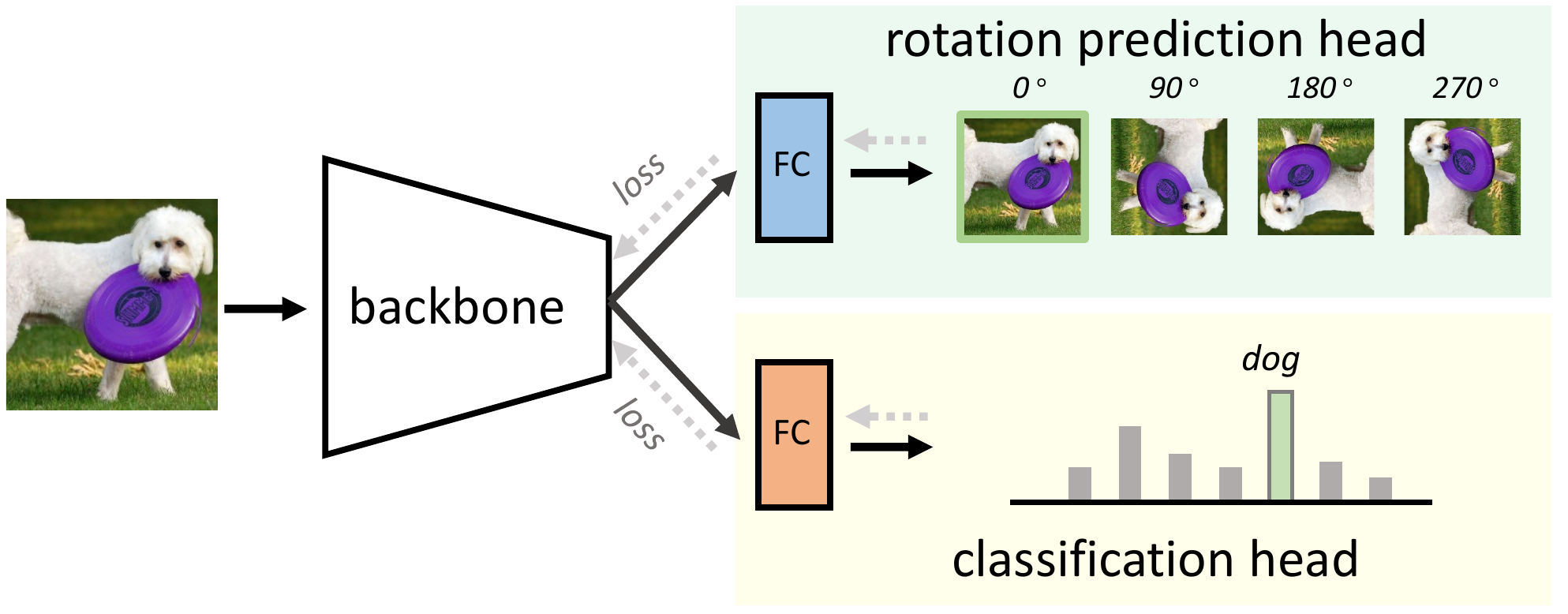}
\caption{Our multi-task network structure for studying the statistical correlation between the accuracy of rotation prediction and classification under different test sets. The network contains a shared backbone (\emph{e.g.}, ResNet-18) for the two tasks, a fully-connected layer for 4-way rotation prediction, and another fully-connected layer for $K$-way classification ($K$ is the number of classes). The shared backbone receives losses from both tasks.}
\label{fig:2}
\end{center}
\end{figure}

\subsubsection{Network Description} \label{multi-task net}
We learn a multi-task neural network for semantic classification along with an auxiliary self-supervised task. The self-supervised task \emph{should} 
1) introduce no learning complexity for the main classification, 
2) require minimal network structure change, and 
3) not degrade classification accuracy.
We choose \emph{rotation prediction} \cite{gidaris2018unsupervised} as the auxiliary task because it meets the above requirements. In the experiment, we observe using Jigsaw also has a strong correlation, but it decreases the classification task accuracy.

\textbf{Rotation prediction} is an effective pretext task initially designed for unsupervised representation learning \cite{gidaris2018unsupervised}. It is shown that predicting rotation degrees helps learn effective features for various downstream tasks, such as recognition, detection and segmentation.

Following the practice in \cite{gidaris2018unsupervised}, we define the set of rotation transformations $G= \{R_{r}(\bm{x})$\}, where $R_{r}$ is rotation transformation and $r\in\{0 \degree, 90 \degree, 180 \degree, 270 \degree \}$.
Given an image, the goal of rotation prediction is to tell which one of the four rotations this image undergoes; it can be formulated as a 4-way classification task.

\textbf{Network structure.} We use a convolutional neural network (ConvNet), \eg, ResNet-50. We only add one fully-connected (FC) layer for rotation prediction and do not modify any other component. The overall structure is very simple, shown in Fig. \ref{fig:2}. The ConvNet has three parts: a shared backbone parameterized by $\bm{\theta}_{s}$, a FC layer for classification parameterized by $\bm{\theta}_{c}$, and another FC layer for rotation prediction parameterized by $\bm{\theta}_{r}$. 

\textbf{Loss function.} Our network is a kind of multi-task model that jointly learns the self-supervised pretext task and semantic classification task. Specifically, the model is trained to simultaneously 1) estimate the geometric transformation applied to an image and 2) classify the semantic category of this image. For rotation prediction, its loss function is written as,
\begin{equation}\small
\mathcal{L}_{\textup{rot}} = \frac{1}{4} \left[ \sum_{  r \in \{ 0^{\circ}, 90^{\circ}, 180^{\circ}, 270^{\circ} \} } \mathcal{L}_{\textup{CE}}(\bm{y}_{r}, \bm{\theta}_{r}(\bm\theta_{s}(R_{r}(\bm{x})))\right],
\end{equation}
where $\bm{y}_{r}$ is the one-hot label of $ r\in\{ 0^{\circ}, 90^{\circ}, 180^{\circ}, 270^{\circ} \}$.

For semantic classification, the loss function is:
\begin{equation}
\mathcal{L}_{\textup{cls}} = \mathcal{L}_{\textup{CE}}(\bm{y}_{c}, \bm{\theta}_{c}(\bm{\theta}_{s}(\bm{x}))),
\end{equation}
where $\bm{y}_{c}$ is the one-hot class label of image $\bm{x}$. The overall loss function of our multi-task network is: $\mathcal{L} = \mathcal{L}_{\textup{cls}} + \mathcal{L}_{\textup{rot}}$.

\textbf{Network inference.} 
\textcolor{black}{After training is complete, we test the multi-task network on unlabeled test images. For the classification task, the network predicts the semantic class for each test image.
{For rotation prediction task, we apply \emph{all} four rotation transformations $G$ to each test image. The network predicts rotations of all transformed images, from which the accuracy of rotation prediction is calculated.}}

\begin{figure*}[t]
\begin{center}
\includegraphics[width=1.0\linewidth]{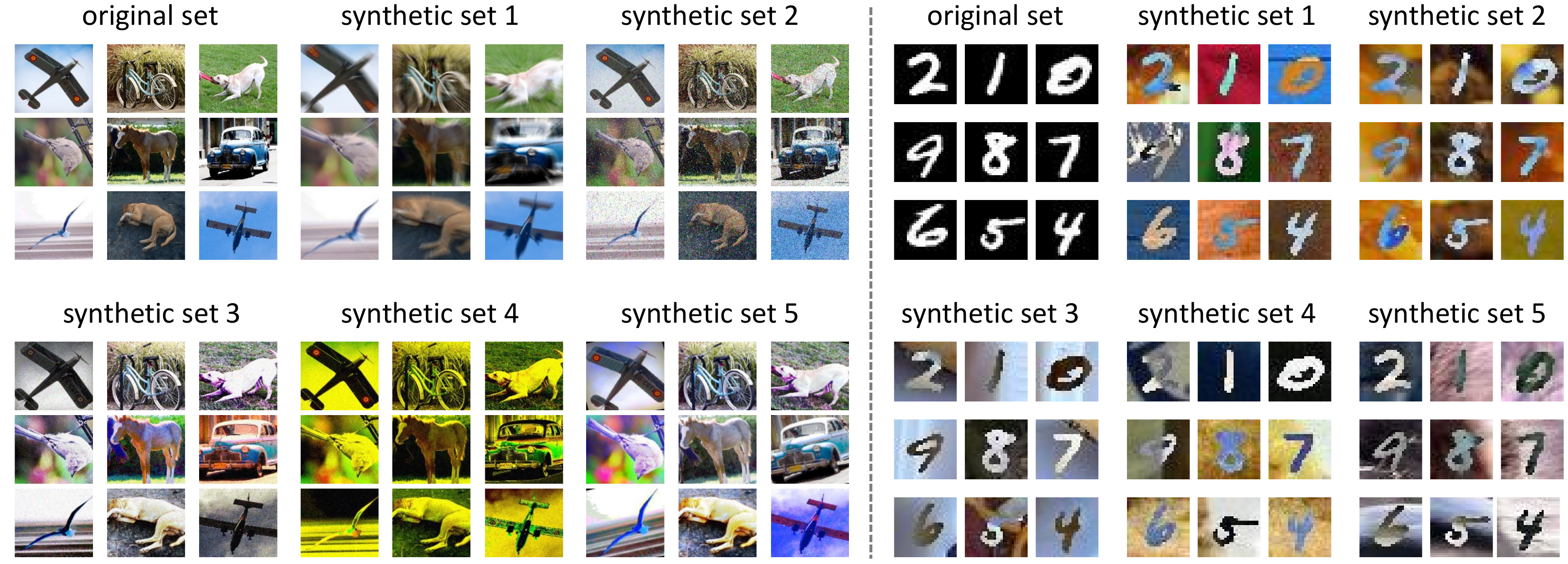}
\caption{Examples of synthetic datasets used in the correlation study. \textbf{Left:}  datasets under the COCO setup. Given an original set (labeled, top left), we use geometric and photometric transforms to synthesize more datasets. For example, we adopt corruption methods \emph{zoom blur} and \emph{impulse noise} to generate ``synthetic set 1'' and ``synthetic set 2'', respectively. We combine several image transformations (\eg, sharpness, ColorTemperature) to produce other synthetic sets.
\textbf{Right:} datasets under the MNIST setup. We change the image background of the original set to synthesize various datasets. 
The synthetic datasets exhibit distinct visual characteristics, and yet inherit the foreground objects from images in the original set, and thus are fully labeled.
}\label{fig:3}
\end{center}
\end{figure*}

\subsubsection{Dataset Synthesis} \label{classifers}
With the network introduced above, we now study the accuracy relationship between rotation prediction and semantic classification under various testing environments. 

To this end, we need many test sets that 1) have various distributions, 2) contain the same classes with the training set, and 3) have sufficient samples. However, there are very few datasets meeting these requirements. Following \cite{deng2020labels}, we synthesize test sets by using various geometric and photometric transforms on top of 
MNIST, CIFAR-10 \cite{krizhevsky2009learning} and COCO \cite{lin2014microsoft}, creating the corresponding setups. 

\textbf{MNIST setup.} MNIST contains 50K training image and 10K test image from 10 categories. We use MNIST training set to train a ConvNet based on LeNet-5. 
To generate various test sets, we change the backgrounds of MNIST test set (named \emph{original set}). Specifically, we use COCO training images as backgrounds and synthesize 1,000 test sets. For all the setups in this section, because the transformations do not change the main image content, the synthesized test sets \emph{inherit the semantic labels} from the original set. 

\textbf{CIFAR-10 setup.} We use DenseNet-40-12 (40 layers with growth rate 12) \cite{huang2017densely} as the backbone. We train our network on CIFAR-10 training set (50K images), and test it on CIFAR-10-C \cite{hendrycks2019benchmarking}. CIFAR-10-C was proposed to study the model robustness to common corruptions. It contains 15 types of corruption and each corruption type comes at 5 levels of severity, resulting in 75 corruption datasets. These corruptions include noise, blur, weather, and digital categories. We refer readers to \cite{hendrycks2019benchmarking} for more details.
To synthesize test sets, we apply image transformations. For each dataset, we randomly select three transformations from \{Sharpness, Equalize, ColorTemperature, Solarize, Autocontrast, Brightness, Rotate\}. Then, we apply three transformations with random magnitudes on images. This practice generates another 425 datasets. 

\textbf{COCO setup.} We use 12 classes, \emph{i.e.}, aeroplane, bike, bird, boat, bottle, bus, car, dog, horse, monitor, motorbike, and person, following \cite{peng2017visda}. We build the training and test sets for the classification task based on COCO training and validation sets, respectively. Each class has 600 images. We also use the 15 common corruptions on the test set and synthesize 75 corrupted test sets. Moreover, we randomly use three transformations to generate another 425 datasets. Examples of synthetic sets are shown in Fig. \ref{fig:3}.

\begin{figure*}[t]
\begin{center}
\includegraphics[width=1.0\linewidth]{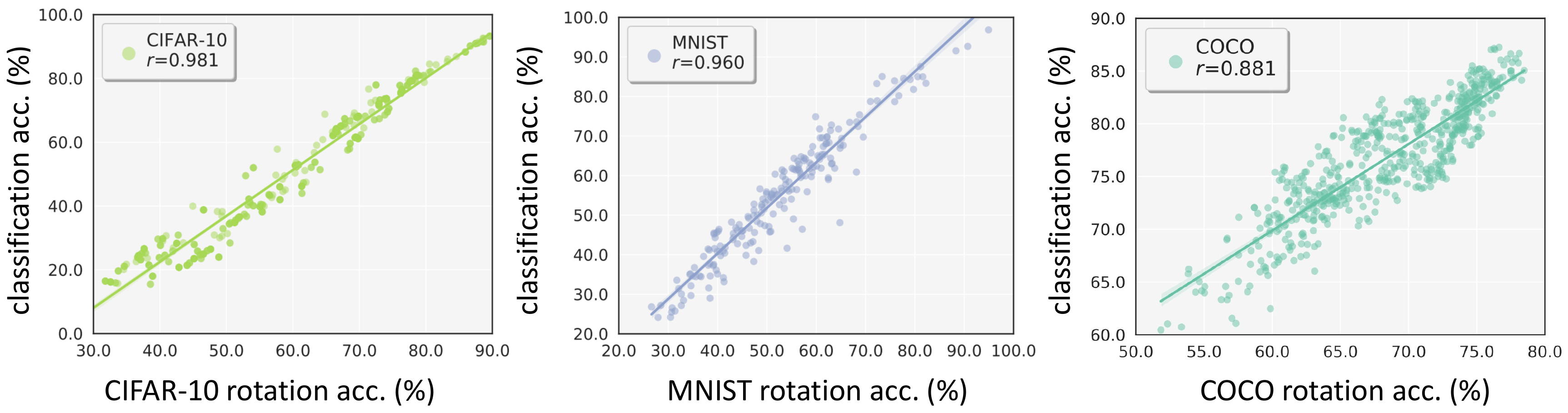}
\vspace{-1em}
\caption{
Correlation between classifier accuracy and rotation prediction accuracy on three setups. From left to right: CIFAR-10 with DenseNet-40-12, MNIST with LeNet-5, and COCO with ResNet-50, respectively. Each point in the figure represents a synthetic test set. Across the three setups, we consistently observe a strong linear relationship (Pearson Correlation $r > 0.88$) between the two accuracy numbers. 
The straight lines are calculated by robust linear regression \cite{huber2004robust}.
}\label{fig:4}
\end{center}
\end{figure*}

\subsubsection{Empirical Analysis}
For each setup introduced above, we compute the accuracy numbers of semantic classification and 4-way rotation prediction on every synthetic test set. We empirically study the statistical relationship between the two tasks' performance using Pearson's Correlation Coefficient ($r$). 

\textbf{Linear correlation.} In Fig.~\ref{fig:4}, we plot semantic classification accuracy against rotation prediction accuracy on these synthetic test sets. 
We have an interesting observation: \emph{given a multi-task model, when the testing environment undergoes changes (i.e., testing on different synthesized test sets), the performance of semantic classification has a {linear} relationship with that of rotation prediction.}
This observation is consistent across three different setups. For example, on the CIFAR-10 and COCO, we find the Pearson's Correlation Coefficient ($r$) is 0.977 and 0.883, respectively, indicating strong linearity. In other words, if the multi-task network is good at predicting rotations, it is most likely to achieve good object recognition accuracy under the same environment, and vice versa.

\subsection{Estimation with Linear Regression}\label{linear_reg}
Based on the strong linear relationship between classification and rotation prediction under changing environments, we train a linear regression model to predict classification accuracy. Specifically, we use $N$ synthetic sets as training samples. The $i$-th ($i=1,2,...,N$) synthetic set is denoted as $(a_i^{rot}, a_i^{cls})$, where $a_i^{rot}$ and $a_i^{cls}$ denote the rotation prediction accuracy and semantic classification accuracy on this set, respectively. The regression model is: 
\begin{equation}\label{linear}
    a^{cls} = w_1 a^{rot} + w_0,
\end{equation}
where $w_1, w_0 \in \mathbb{R}$ are linear regression parameters.
In practice, we use robust linear regression \cite{huber2004robust} with Huber loss to learn the regression model. 
In Section \ref{sec:experiment}, we will present results of using the learned linear regression models for accuracy prediction on unseen test sets.

\section{Experiment}\label{sec:experiment}
\subsection{Task Setting and Baseline}
\setlength{\tabcolsep}{5pt}
\begin{table*}[t]
    \caption{Results of classifier accuracy prediction using various methods on three groups of unseen test sets: 1) SVHN and USPS; 2) CIFAR-10.1; 3) Pascal, Caltech and ImageNet. The training set for each group is MNIST, CIFAR-10, and COCO, respectively.
For each test set, we report the estimated classification accuracy (\%) and the ground-truth recognition accuracy (\%). RMSE (\%) is reported to measure the estimation precision. For ``prediction'' and ``entropy'' methods, images with high prediction scores (greater than $\tau_1$) or low entropy scores (less than $\tau_2$) are regarded as being classified correctly, respectively.}
    \begin{center}
    \begin{tabular}{l|cc|c|c|c|ccc|c}
  			\toprule
     Train Set&\multicolumn{3}{c|}{MNIST}&\multicolumn{2}{c|}{CIFAR-10}&\multicolumn{4}{c}{COCO}\\
    \hline
    Unseen Test Set & SVHN &  USPS & RMSE$\downarrow$&  CIFAR-10.1 & RMSE$\downarrow$ & Caltech & Pascal & ImageNet & RMSE$\downarrow$\\
    \hline
    Ground-truth Accuracy & 23.06 & 65.52 & - & 88.15 & - & 92.61 & 86.43 & 87.83 &-\\
    \hline
    Prediction ($\tau_1 =0.8$) & 33.64 & 44.34& 16.74 & 91.15 & 3.00 & 89.36 & 83.98 & 85.17 & 2.81\\
    Prediction ($\tau_1 =0.9$) & 22.07 & 30.39 & 24.85 & 86.85 & 1.30 & 84.30 & 78.00 & 79.83 & 8.25\\
    \hline
    Entropy ($\tau_2 =0.2$) & 26.63& 33.23 & 22.97& 89.20 & 1.05 & 86.80 & 80.14 & 82.50 & 5.82\\
    Entropy ($\tau_2 =0.3$) & 40.35& 46.87 & 17.98& 93.80 & 5.65 & 92.49 & 86.21 & 88.50 & 0.41\\
    \hline
    \rowcolor{weijian} Linear Regression & 24.84  & 53.10 &8.87 & 91.89 & 3.74 & 90.70 & 89.29 & 90.98 & 2.68\\
			\bottomrule
    \end{tabular}
    \end{center}
    \label{tab:estimate}
\end{table*}
\textbf{Task setting.} For the problem of unsupervised classifier evaluation, we experiment on two semantic classification settings, \emph{i.e.,} digit classification and natural image classification. 
In each setting, we estimate the semantic classification accuracy on \emph{unseen test sets}.

For digit classification, we test LeNet-5 (trained on MNIST) on two unseen real datasets, \ie, USPS \cite{hull1994database} and SVHN \cite{netzer2011reading}, both with 10 classes. For natural image classification, we use ResNet-50 trained on COCO and estimate classification accuracy on three datasets: PASCAL \cite{everingham2007pascal}, Caltech \cite{griffinHolubPerona} and ImageNet \cite{deng2009imagenet}, all with 12 classes. 
In addition, we use CIFAR-10.1 \cite{recht2018cifar} for CIFAR-10 setup: we estimate semantic classification accuracy of DenseNet-40-12 on this unseen test set through ConvNet trained on CIFAR-10.

\textbf{Metric.} To evaluate the correctness of performance estimation, we use root mean squared error (RMSE). RMSE measures
the average squared distance between the estimated accuracy and
ground-truth accuracy. Small RMSE corresponds to good estimation and vice versa.

\textbf{Baselines.} There are very few studies on the unsupervised accuracy estimation problem. Thus, we introduce two intuitive methods for comparison, which are inspired by the uncertainty estimation methods \cite{hendrycks2016baseline,kendall2017uncertainties,liang2017enhancing}.

\emph{Prediction score based.} Given an input image, its score is obtained as the maximum softmax output. If the score is greater than a threshold $\tau_1 \in [0,1]$, this image is regarded as being correctly classified.

\emph{Entropy score based.} The score is given by the entropy of softmax outputs, normalized by $log(K)$, where $K$ is the number of classes. Predictions with scores less than a threshold $\tau_2 \in [0, 1]$ are considered correct. 

\subsection{Experimental Results}
In Table~\ref{tab:estimate}, we report the performance of our linear regression models on the six real-world test sets described above. We have the following observations.

\textbf{Score-based methods are sensitive to thresholds.} As shown in Table~\ref{tab:estimate}, the two intuitive score-based methods could produce good prediction results when using proper thresholds. However, they are sensitive to the threshold value. For example, using $\tau_1=0.9$ and $\tau_1=0.8$ will lead to a 5.44\% difference in RMSE under the COCO setup. 
Furthermore, different datasets need different thresholds. For instance, for the prediction score based method, $\tau_1=0.9$ is suitable at CIFAR-10.1, but is not good for three natural image classification datasets.
Notably, it is hard to select the optimal threshold because 1) test labels are unavailable and 2) the testing environment keeps changing. That said, designing a strategy to select thresholds might be a potential way to improve the practicability of two intuitive methods.

\textbf{Our linear regression model is feasible to predict classifier performance.} The linear regression model (Section~\ref{linear_reg}) based on rotation prediction accuracy could achieve promising predictions across three setups, shown in Table~\ref{tab:estimate}.
For example, when predicting the accuracy of a classifier on three test sets under the COCO setup, our regression model has a 2.68\% RMSE. In addition, it estimates the accuracy to be 91.89\% on CIFAR-10.1 (the ground truth is 88.15\%). 
Since the training samples (synthetic datasets) of the regression model are diverse enough (\eg, diverse distributions and various visual characteristics), it can learn to achieve reasonably accurate estimations on unseen test sets.

\begin{figure}[t]
\begin{center}
\includegraphics[width=1.0\linewidth]{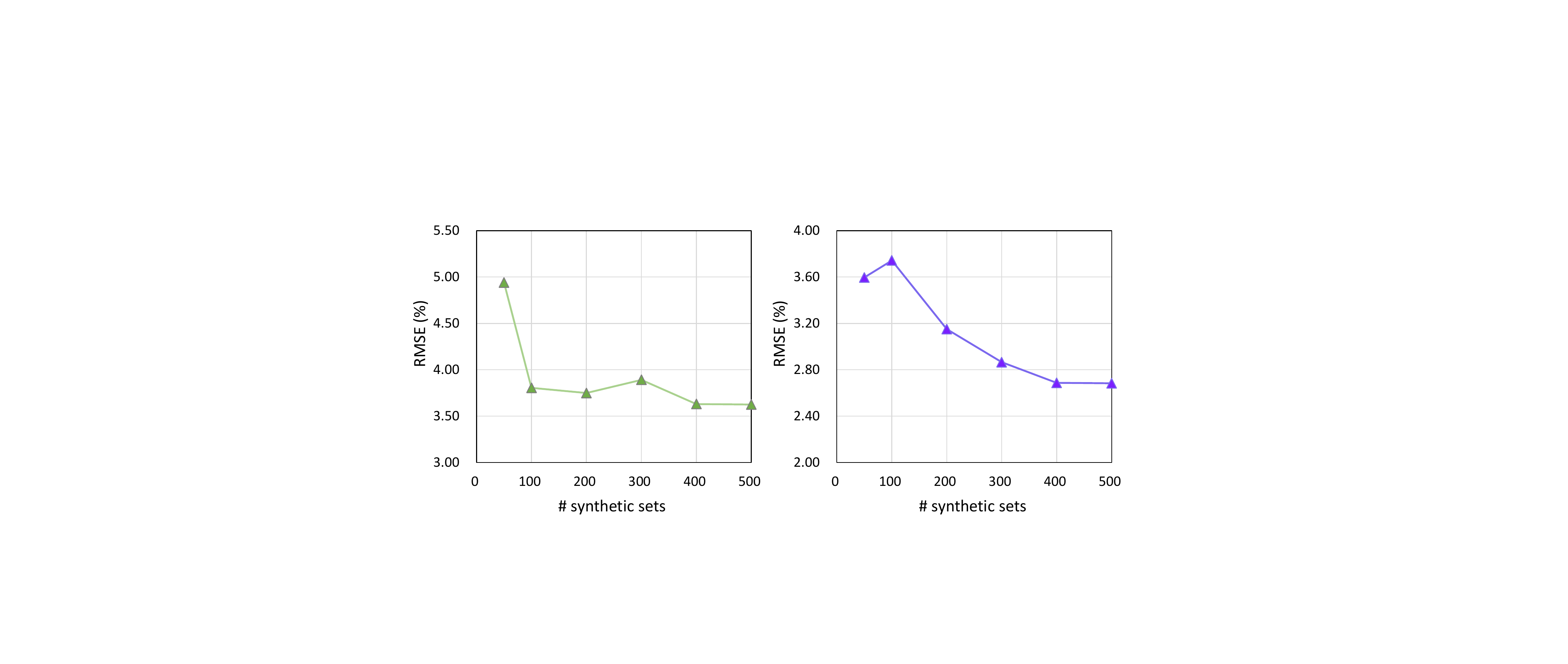}
\vspace{-1.5em}
\caption{
Impact of the number of training samples (synthetic sets) for linear regression. \textbf{Left:} RMSE (\%) on CIFAR10.1. \textbf{Right:} RMSE (\%) on Caltech, Pascal, and ImageNet.
We observe that the regression model becomes stable when using more than 200 training samples.
}
\label{fig:sample_size}
\end{center}
\vspace{-10pt}
\end{figure}

\textbf{Impact of the number of training samples (synthetic sets).} 
In Fig.~\ref{fig:sample_size}, we present the RMSE (\%) obtained by linear regressors trained using different numbers of synthetic sets. Since the linear regression model only has two parameters, it does not require too many training samples \cite{huber2004robust}. In our problem, when using more than 200 synthetic sets, RMSE is relatively stable. In practice, we speculate that using sufficient and diverse synthetic sets is necessary to improve regression robustness.
\begin{figure}[t]
\begin{center}
\includegraphics[width=1.0\linewidth]{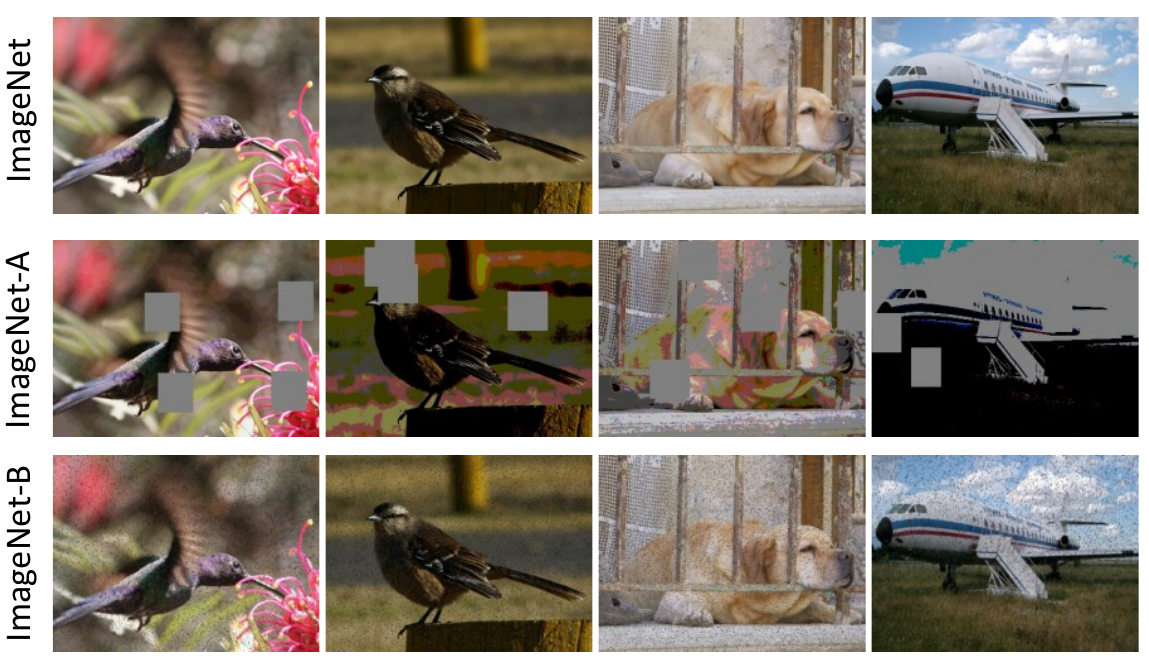}
\caption{Visual examples of transformed natural ImageNet images. For ImageNet-A, we use \emph{Cutout} and \emph{Posterize} to edit original images. For ImageNet-B, we use \emph{Pepper} and \emph{FilterSmooth} to introduce visual changes.
}
\label{fig:6}
\end{center}
\vspace{-10pt}
\end{figure}

\textbf{Robustness of linear regressor on transformed test sets.} We apply various image transformations to edit three natural image datasets (ImageNet, Pascal and Caltach), and test the regression model on them.  Specifically, we use \emph{Cutout} and \emph{Posterize} to produce ImageNet-A, Pascal-A and Caltach-A, and use \emph{Pepper} and \emph{FilterSmooth} to produce ImageNet-B, Pascal-B and Caltach-B. Visual examples of these datasets are shown in Fig.~\ref{fig:6}. 
Note that these transformations \emph{were not used} when synthesizing datasets to train the regression model. 
Thus, testing on them allows us to understand the robustness of the regression model. 

\begin{figure}[t]
\begin{center}
\includegraphics[width=1.0\linewidth]{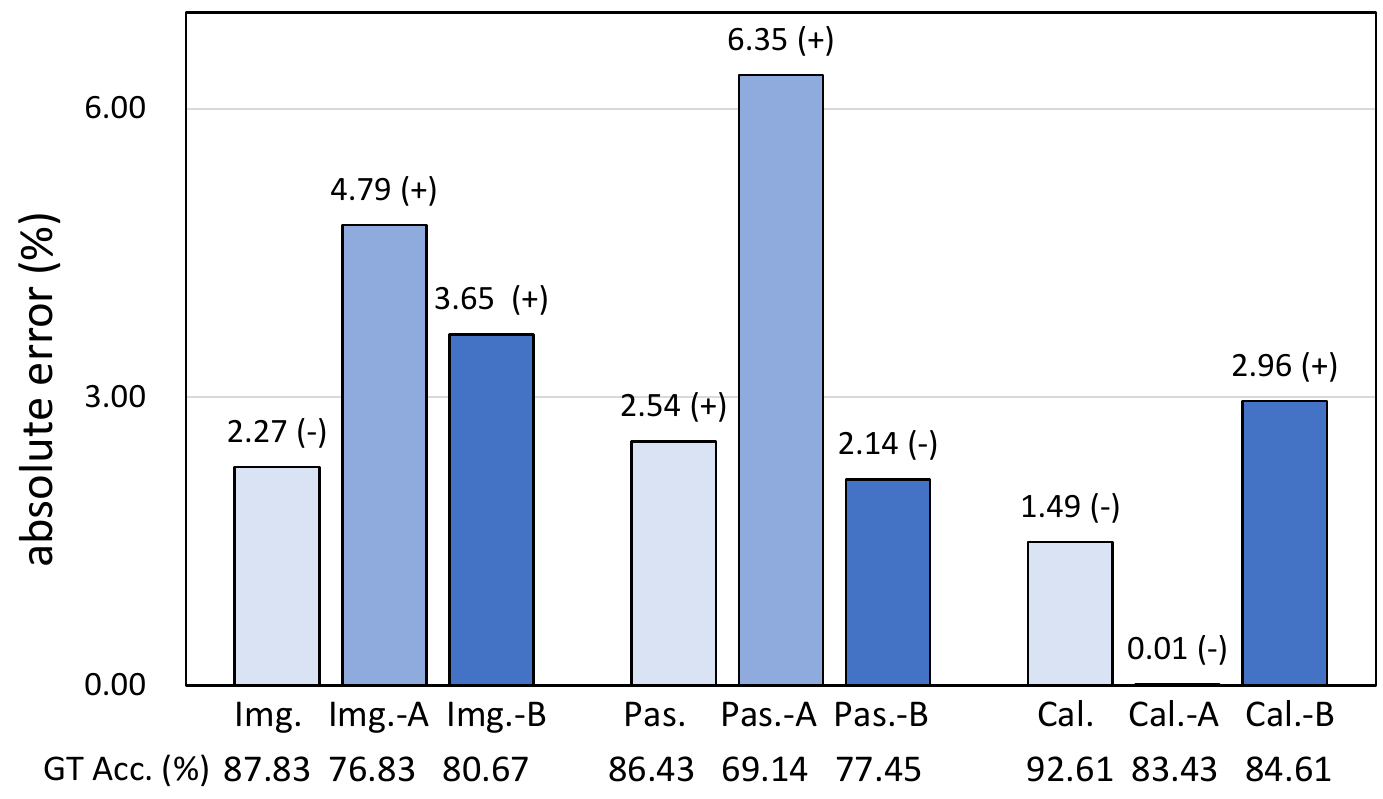}
\caption{Linear regression on transformed test sets (ImageNet, Pascal, and Caltech). The transformed datasets are denoted by ``-A" and ``-B".
The absolute error (\%) between estimated classier accuracy and the ground truth accuracy (\%) (shown below each dataset) is reported. (-) / (+) denotes the predicted accuracy is lower and higher than the ground-truth accuracy, respectively.
}
\label{fig:7}
\end{center}
\end{figure}

We have the following observations in Fig.~\ref{fig:7}. First, the transformations introduce extra visual changes, so the classifier accuracy decreases on the edited dataset. Second, the regression model still makes good estimated results on the edited test sets. For example, it makes good predictions on Caltech-A and Caltech-B, as the absolute errors are 0.01\% and 2.95\%, respectively. 
\textcolor{black}{Although the absolute error on Pascal-A is 6.35\%, the estimated accuracy still helps us gain a general understanding of classifier behaviour on this dataset.}
We think the regression precision can be further improved by using other information, such as distribution shift \cite{deng2020labels}. 
Moreover, \textcolor{black}{if we are provided with a small number of labeled test images, it could be helpful for adjusting regression predictions.}

\begin{figure}[t]
\begin{center}
\includegraphics[width=0.88\linewidth]{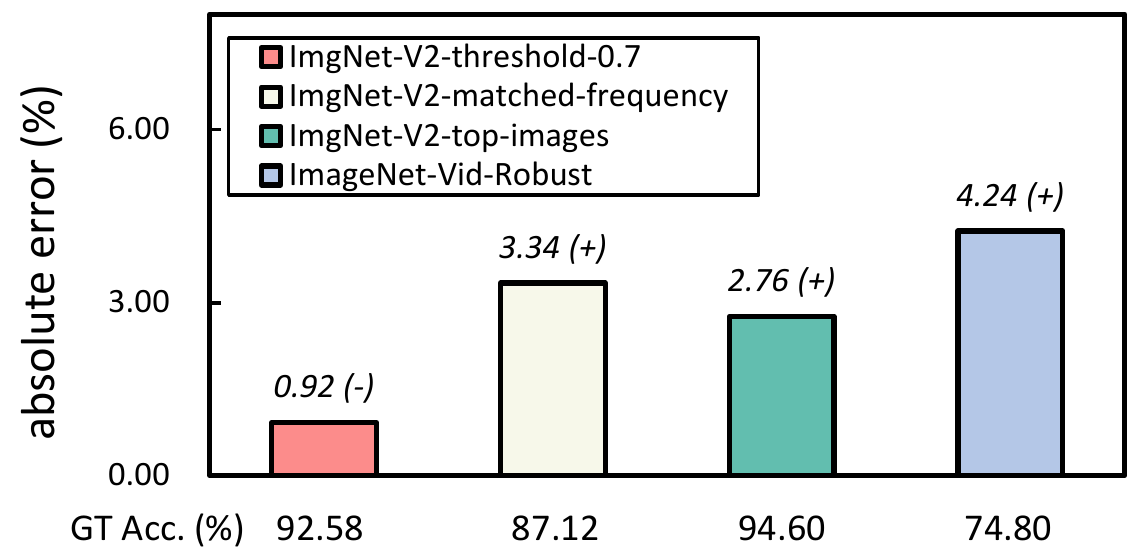}
\caption{The absolute error (\%) of predictions on \emph{four natural} test sets under COCO setup. The ground truth (\%) is shown under each dataset. (-) / (+) means the predicted accuracy is lower / higher than the ground-truth accuracy, respectively.
}
\label{fig:8}
\end{center}
\end{figure}

\textbf{Generalization ability of the regression model.}  To further validate this, we test on \emph{four natural test sets} with different types of domain shifts \cite{taori2020measuring} and report estimations in Fig. \ref{fig:8}. We observe that our method overall gains good estimations, and its the average RMSE of the four datasets is 3.07\%. 

Based on the above experimental analysis, we think self-supervised rotation prediction provides a feasible way to predict classifier accuracy on varying testing environments without annotated category labels.

\section{Discussion}
\textbf{When trained in a multi-task ConvNet, does rotation prediction affect semantic classification accuracy?} To answer this question, we conduct experiments to study whether classification is compromised by the introduction of the rotation prediction head.
In Table~\ref{tab2}, we use three types of networks (ResNet-26, DenseNet-40-12, and VGG-19). We observe that auxiliary rotation prediction task \emph{does not} degrade main classification on the CIFAR-10 test set.

In fact, rotation prediction requires the neural network to capture the rotation-related representations of an image (\eg, particular structures and salient object-level patterns), which are also useful for objection recognition \cite{gidaris2018unsupervised}.
Moreover, while rotation prediction does not improve accuracy on original or clean testing image, it is reported to improve classifier robustness to input corruptions \cite{hendrycks2019using}. In addition, recent works \cite{sun2019unsupervised,bucci2021self} have shown that using rotation prediction for training can improve the domain adaptation ability of the classifier.
Based on the above discussion, introducing auxiliary rotation prediction task \emph{will not} negatively impact the main classification task.

\setlength{\tabcolsep}{7pt}
\begin{table}[t]
\caption{Impact of the auxiliary rotation prediction (Rot.) task on the main classification task. We conduct experiment on CIFAR-10 with ResNet26, Dense40-12, and VGG19. Top-1 accuracy (\%) is reported. We show that the rotation prediction task \emph{does not} affect the classifier accuracy.}
\small
\begin{center}
\begin{tabular}{c|ccc}
		\toprule
{\multirow{2}{*}{Train w/ Rot.?}}&\multicolumn{3}{c}{CIFAR-10}\\
\cline{2-4}
&ResNet26&Dense40&VGG19\\
\hline
&92.86$\pm$0.10&94.65$\pm$0.15&92.31$\pm$0.13\\
\hline
\checkmark&92.84$\pm$0.10&94.75$\pm$0.15&92.51$\pm$0.20\\
	\bottomrule 
\end{tabular}
\end{center}
\label{tab2}
\end{table}
\begin{figure}[t]
\begin{center}
\includegraphics[width=1.0\linewidth]{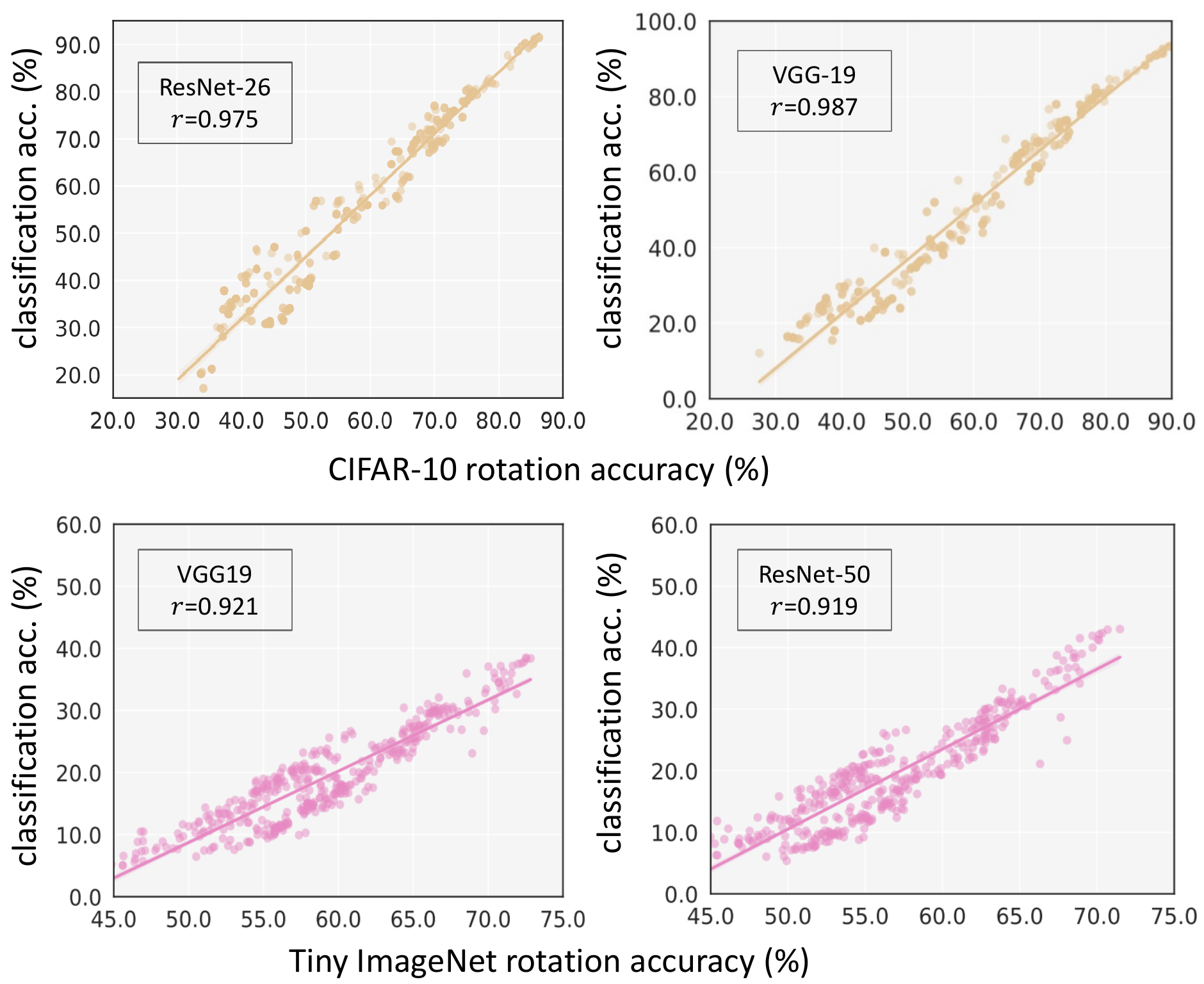}
\caption{Correlation between semantic classification accuracy and rotation prediction accuracy when using difference types of network structures. On CIFAR-10, we use ResNet-26 and VGG-19. The Pearson Correlation $r$ is 0.975 and 0.987 with ResNet-26 and VGGNet-19, respectively.
On Tiny ImageNet, the correlation $r$ is 0.921 and 0.919 using VGG-19 and ResNet-50, respectively.
}
\label{fig:5}
\end{center}
\end{figure}

\textbf{Does the linear relationship hold when using other backbone structures?} 
To study this, we use different types of backbones for correlation study and report results in Fig.~\ref{fig:5}. 
We also include Tiny ImageNet for the correlation study. It has 200 classes, and each class has 500 training images. We follow the COCO setup to synthesize 500 datasets.

For CIFAR-10, when using ResNet-26 and VGGNet-19 as backbones, we observe a strong linear correlation between the two task performance. Specifically, Pearson's Correlation ($r$) is 0.975 and 0.987 with ResNet-26 and VGGNet-19, respectively.
When using VGG-19 and ResNet-50 on Tiny ImageNet, we also observe the correlation ($r$) is 0.921 and 0.919, respectively. It also indicates the strong linear correlation between two task accuracy on varying test sets is maintained when using different backbones.

In addition, we show the correlation values on CIFAR-10 setup using different backbones in Table~\ref{tab3}. In the experiment, we use five backbones: VGG11, VGG19, ResNet26, ResNet44, and Dense40-12. 
\textcolor{black}{We have two observations. First, the correlation values are very high ($r>0.975$) when using different backbones. Another interesting finding is: the performance of two tasks is varying with different backbones, but their correlation on different test sets remains at the same level: \emph{the ``mean $\pm$ std" of Pearson's Correlation ($r$) is $0.983 \pm 0.005$}}. 

\setlength{\tabcolsep}{3pt}
\begin{table}[t]
\caption{
Linear correlation ($r$) on CIFAR-10 with different backbone structures, including VGG11, ResNet26, VGG19, ResNet44, and Dense40-12. We also include the accuracy (\%) of classification (Class. Acc) and rotation prediction (Rot. Acc) on the CIFAR-10 test set. Using different backbones, the two task performance is varying, but their correlation remains at the same level: {the ``mean $\pm$ std" of correlation ($r$) is $0.983 \pm 0.005$}.
}
\begin{center}
\small
\begin{tabular}{c|c|c|c|c|c}
	\toprule
              & VGG11   & VGG19 & ResNet26 & ResNet44 & Dense40 \\ \hline		
Class. Acc.   & 92.53   & 92.51 & 92.84 & 93.73    & 94.75         \\ \hline
Rot. Acc.     & 91.32   & 92.07 & 87.84 & 88.81    & 91.28         \\ \hline
Cor. ($r$) & 0.990   & 0.987 & 0.975 & 0.981   & 0.981          \\ 
\bottomrule
\end{tabular}
\end{center}
\vspace{-10pt}
\label{tab3}
\end{table}

\textbf{Impact of the number of categories on correlation.} For rotation prediction, it distinguishes between four different classes, but for image classification, the number of categories can be much greater. In Fig. \ref{fig:4} and Fig. \ref{fig:5}, we observe the correlation is high ($r >0.88$) across different datasets with various numbers of categories. For example, the correlation is  $r>0.90$ on Tiny-ImageNet with 200 categories. 

We further study on CIFAR-100 setup and report results in Table \ref{tab:4}. We follow the CIFAR-10 setup to synthesize 500 datasets for CIFAR-100 setup. Although the correlation $r$ is lower than that on CIFAR-10 (0.918 vs. o.975 with ResNet26), the value $r=0.918$ still indicates strong linear correlation under the CIFAR-100 setup. Based on the above analysis, we speculate when the number of categories is huge (\eg, 10K \cite{deng2010does}), the correlation $r$ might decrease but it will still have a high value. 

\textcolor{black}{\textbf{Why is rotation prediction performance closely correlated with classification performance?} Our multi-task network jointly learns both tasks. By doing so, they are coupled together and share the feature representations. When testing on images from the novel distribution, the dataset shift will make the presentations not suitable for the test set. Thus, the network performance on both tasks drops simultaneously. 
That is, different test sets introduce different dataset shifts, leading to varying degrees of performance drop. 
This might be the reason why their performance shows a positive correlation trend on varying test domains.}

\setlength{\tabcolsep}{6pt}
\begin{table}[]
\caption{Effect of the number of semantic categories on the correlation. Although the rank correlation ($\rho$) on CIFAR-100 is slightly lower than that on CIFAR-10, it still indicates that the two tasks have a strong linear correlation ($\rho>0.9$).
The accuracy (\%) of classification (Class. Acc) and rotation prediction (Rot. Acc) on CIFAR-100 test set is also reported. }
\begin{center}
\small
\begin{tabular}{c|c|ccc}
	\toprule
\multirow{2}{*}{Backbone}   
         & \multicolumn{1}{c|}{CIFAR-10} & \multicolumn{3}{c}{CIFAR-100} \\
          \cline{2-5}
         & Cor. ($r$)       &  Cor. ($r$)         &  Class Acc.  & Rot. Acc.    \\ \hline
ResNet26 & 0.975               &     0.918              &   69.31      &   73.18      \\ \hline
ResNet44 & 0.981               &     0.910              &   71.38      &   75.60      \\ \hline
Dense40 & 0.981                &     0.950              &   74.55      &   75.20     \\ \bottomrule
\end{tabular}
\end{center}
\vspace{-10pt}
\label{tab:4}
\end{table}

\textbf{Difference from those using self-supervision for classification \cite{sun2020test,hendrycks2019using,carlucci2019domain,zhai2019s4l}.} These works present that the \emph{self-supervised} task can help \emph{supervised} classification task in different perspectives, such as robustness \cite{hendrycks2019using} and domain adaptive ability \cite{sun2020test}. This work differs significantly: 
our work does not aim to improve supervised classification by using self-supervised tasks. Rather we offer a new and interesting perspective into the relationship between the two tasks, a finding that allows estimation of generalization performance for semantic classification models from easily obtained self-supervised rotation prediction results.

\textbf{Limitation.} We might encounter corner cases (\emph{e.g.}, balls and airplanes) where they are even impossible for a human to tell the image rotation angle. The proposed method, because of the intrinsic limitation of rotation prediction, would fail in such cases. In fact, as an underlying assumption, the self-supervised task is required to be well-defined and non-trivial \cite{sun2020test}. Moreover, we might encounter images from unseen classes in the open world. To improve the estimation under this scenario, a feasible solution is using out-of-distribution detection techniques \cite{hsu2020generalized,hendrycks2016baseline,lee2018simple,liang2017enhancing} to detect and reject such novel images.

\textbf{Feasibility of other self-supervised tasks than rotation prediction.} There are many other self-supervised tasks, such as jigsaw puzzles \cite{noroozi2016unsupervised} and colorization \cite{larsson2017colorization}. Since our network jointly learns semantic classification and pretext tasks, we select the pretext task based on the criteria listed at the beginning of Section \ref{multi-task net}.
In the preliminary experiment, we found that \emph{colorization} discards the color information of images and thus makes the learning of semantic classification unstable. In addition, we observe that \emph{Jigsaw} and semantic classification exhibit a linear correlation ($r > 0.950$) under the CIFAR-10 setup (Table \ref{tab5}). Nevertheless, in our multi-task network, Jigsaw compromises the classification accuracy by about $2\%$, while rotation prediction does not affect semantic classification (shown in Table \ref{tab2}). 
Moreover, our network needs to learn the semantic classification and pretext task together. It would be interesting to study other potential ways to use pretext tasks for estimating classifier accuracy.

\setlength{\tabcolsep}{6pt}
\begin{table}[]
\caption{The extent of linear correlation ($r$) on CIFAR-10 when using Jigsaw or rotation prediction as the pretext task. The accuracy (\%) of classification (Class. Acc) on the CIFAR-10 test set is also reported. Jigsaw and semantic classification also have a strong linear correlation ($r>0.95$). However, it degrades the classification accuracy when trained in the multi-task network. 
}
\vspace{1.5em}
\begin{center}
\small
\begin{tabular}{c|cc|cc}
	\toprule
\multirow{2}{*}{Backbone}   
         & \multicolumn{2}{c}{Rotation} & \multicolumn{2}{c}{Jigsaw}\\
          \cline{2-5}
          &  Cor. ($r$)       &  Class Acc.  & Cor. ($r$)  & Class Acc.  \\ \hline
ResNet26  &     0.975         &   92.84      &   0.953     & 90.90 \\ \hline
ResNet44  &     0.981         &   93.73      &   0.958     & 91.81 \\
\bottomrule
\end{tabular}
\end{center}
\label{tab5}
\end{table}

\section{Conclusion}
In this paper, we try to estimate classifier accuracy under varying testing environments without human-annotated labels. We report a strong linear correlation ($r>0.88$) between semantic classification accuracy and rotation prediction accuracy through empirical studies. This interesting finding makes it feasible to estimate classifier accuracy on unlabeled test sets using rotation prediction performance which can be easily obtained. The extensive experiments provide supports and insights for using such a pretext task to predict the generalization ability of the classifier.

\section*{Acknowledge}
{This work was supported in part by the Australian Research Council (ARC) Discovery Early Career Researcher Award (DE200101283), ARC Centre of Excellence in Robotic Vision (CE140100016), and the ARC Discovery Project (DP210102801). We thank all anonymous reviewers for their constructive comments in improving this paper.}

\bibliography{egbib}
\bibliographystyle{icml2021}

\end{document}